\documentclass[lettersize,journal]{IEEEtran}
\usepackage{amsmath,amsfonts}
\usepackage{algorithmic}
\usepackage{algorithm}
\usepackage{array}
\usepackage[caption=false,font=normalsize,labelfont=sf,textfont=sf]{subfig}
\usepackage{textcomp}
\usepackage{stfloats}
\usepackage{url}
\usepackage{verbatim}
\usepackage{graphicx}
\usepackage{cite}

\usepackage{siunitx}
\usepackage{xcolor}
\usepackage{hhline}
\usepackage{multirow}
\usepackage{csquotes}
\usepackage{float}
\usepackage{makecell}

\definecolor{rev-color}{rgb}{0, 0, 0}
\definecolor{rev-color2}{rgb}{0, 0, 0}
\definecolor{rev-color3}{rgb}{0, 0, 0}

\begin{document}

\title{Latency Adjustable Transformer Encoder for Language Understanding}

\author{Sajjad Kachuee, Mohammad Sharifkhani\\Department of Electrical Engineering\\Sharif University of Technology}

\maketitle

\begin{abstract}
Adjusting the latency, power, and accuracy of natural language understanding models is a desirable objective of an efficient architecture. This paper proposes an efficient Transformer architecture that adjusts the inference computational cost adaptively with a desired inference latency speedup. In fine-tuning phase, the proposed method detects less important hidden sequence elements (word-vectors) and eliminates them in each encoder layer using a proposed Attention Context Contribution (ACC) metric. After the fine-tuning phase, with the novel offline-tuning property, the inference latency of the model can be adjusted in a wide range of inference speedup selections without any further training.
\textcolor{rev-color3}{
Extensive experiments reveal that most word-vectors in higher Transformer layers contribute less to subsequent layers, allowing their removal to improve inference latency. Experimental results on various language understanding, text generation, and instruction tuning tasks and benchmarks demonstrate the approach's effectiveness across diverse datasets, with minimal impact on the input's global context. The technique improves Time-to-First-Token (TTFT) of Llama3 by up to 2.9x, with minor performance drop.} The suggested approach posits that in Large Language Models (LLMs), although the complete network is necessary for training, it can be truncated during the fine-tuning phase.
\end{abstract}

\textcolor{rev-color}{
}

\section{Introduction}
Recently, Transformer \cite{vaswani2017attention} based architectures achieved remarkable success in various Natural Language Processing (NLP) tasks. However, the training difficulty, inference latency, and small size of available datasets are the main concerns with using these models in real applications \cite{strubell2020energy}. Fine-tuning a pre-trained language model on downstream tasks mitigates some of these concerns, including training effort and dataset size. \textcolor{rev-color3}{BERT \cite{devlin2018bert}, XLNet \cite{yang2019xlnet}, GPT \cite{brown2020language}, Llama \cite{touvron2023llama}, and Gemma \cite{team2024gemma} represent a spectrum of pre-trained language models, ranging from earlier developments to the latest advancements. These models vary significantly in size, ranging from hundreds of millions to hundreds of billions of parameters.} Hence, the computational complexity bears \textcolor{rev-color3}{trillions} of Floating Point Operations (FLOPs) for processing a single sample. In some cases, the inference phase computational effort is beyond the ability of resource-limited systems such as edge devices. Therefore, baseline Transformer models cause unacceptable latency and energy usage limitations in such devices.

\textcolor{rev-color3}{
BERT (Bidirectional Encoder Representations from Transformers) and GPT (Generative Pre-trained Transformer) are two landmark architectures in the development of large language models (LLMs). BERT, introduced by Google, uses a bidirectional training approach, allowing it to capture context from both directions of a sentence simultaneously. This makes it particularly effective for tasks like sentence classification, question answering, and named entity recognition, where understanding relationships between words is crucial \cite{devlin2018bert}. On the other hand, GPT, developed by OpenAI, takes an autoregressive approach, generating text word by word in a unidirectional manner. Starting with GPT-1 and evolving through GPT-4, GPT has proven its strength in creative and coherent text generation, conversational AI, and text summarization, making it a highly flexible model for both content generation and interactive applications \cite{radford2018improving, brown2020language}.
}

\textcolor{rev-color3}{
More recent models like Meta's LLaMA (Large Language Model Meta AI) and Gemma are pushing the boundaries of LLM development. LLaMA is designed to be highly efficient in terms of computational resources, offering a smaller model that still performs well on a wide range of natural language processing tasks. Its streamlined design makes it accessible for research and industrial applications where resources might be limited \cite{touvron2023llama}. Gemma, an emerging model family, is expected to further refine the efficiency and adaptability of LLMs, with an emphasis on real-time performance and scalability. Gemma is particularly well-suited for industry-specific applications, aiming to bridge the gap between general-purpose LLMs and more specialized use cases \cite{team2024gemma}. Together, these model families reflect the rapid progression and specialization in the field of large language models, each offering distinct strengths based on the needs of users and developers.
}

\textcolor{rev-color}{
In the realm of Deep Neural Networks (DNNs) optimization, diverse methods have been proposed to enhance speed and reduce power consumption. This is crucial for applications such as extending battery life and enabling powerful tasks on small devices. The approaches can be categorized into pruning, removing unnecessary parts; quantization, simplifying numerical usage; Knowledge Distillation, simplifying based on a complex model; parameter-sharing, using the same information across components; tensor decomposition, breaking down data structures; and sub-quadratic complexity transformers \cite{gupta2022compression}.}

\textcolor{rev-color}{
This study introduces an innovative attention layer designed to enhance the efficiency of inference stages in Transformer-based models with an adjustable approach. The suggested solution involves a sorting and elimination process for hidden sequence elements (word-vectors), resulting in a reduction of effective word-vectors in each layer and a decrease in the number of FLOPs. The primary contributions of this research can be summarized as follows:
}

\begin{itemize}
\item The novel structure and tuning policy are proposed to speed up the inference times with negligible accuracy degradation.
\item A new Attention Context Contribution (ACC) metric is proposed to observe the context contribution in each attention layer. It is shown that the context contribution is decreased at the last encoder layers significantly.
\item An accurate analytical inference speedup estimation is presented for the proposed approach that helps a system designer determine the hyper-parameter selection at the fine-tuning phase.
\item Offline-tuning property that adjusts the inference latency after the fine-tuning phase. This property controls the trade-off between latency and accuracy in a wide range of inference speedup selections without requiring additional training.
\textcolor{rev-color3}{\item The proposed approach can be applied to a variety of Transformer-based architectures, while preserving the global context of the input.}
\end{itemize}

\section{Related Works}\label{Related_Works}

\textcolor{rev-color}{
Pruning-based methodologies have been extensively explored in the existing body of literature, with categorizations based on distinct pruning approaches. In weight pruning methods, the model's weights are systematically pruned according to a predetermined threshold \cite{wiedemann2019compact, jiang2022model, tang2022automatic, ganesh2022slimming}. Head and channel pruning methods extend their scope by entirely excising a channel or head of the network \cite{ma2022small, chen2023three, guan2022dais, peng2022recnas, qian2023hierarchical, liu2022soks, zheng2022model, salehinejad2021edropout, zhang2021structadmm, chen2020dynamical, lin2021network}. Layer and block pruning strategies \cite{li2022stage, zhou2021evolutionary, bian2021subarchitecture}, involve the targeted removal of entire layers or blocks within the network. Token and word-vector pruning methods, adopting a more fine-grained approach, focus on the selective removal of specific tokens or word-vectors within the model \cite{goyal2020power, dai2020funnel, kim2022learned}. This classification clarifies the various approaches taken in the pruning paradigm, each playing a unique role in achieving the overall goal of optimizing networks.
}

\subsection{Weight Pruning}
In \cite{gordon2020compressing}, the magnitude weight pruning method is applied to BERT and shows that the model can be pruned once during pre-training rather than separately for each task without affecting performance. \textcolor{rev-color}{Reweighted Proximal Pruning (RPP) has introduced a pruning method specifically designed for a large-scale language representation model that is much better than typical iterative pruning. This approach yields a sparse version of BERT \cite{guo2019reweighted}. Transformer mixed-length vector pruning (TF-MVP) examines the sparsity patterns in pruned Transformers and introduces the TF-MVP architecture using the mixed-length vector pruning (MVP) technique \cite{yoo2023tf}. While TF-MVP achieves a pruning ratio of approximately 70\%, there is a notable decline in accuracy.} In \cite{louizos2017learning}, the model weights are encouraged to be zero with the L0 regularization term.

\subsection{Head/Channel Pruning}
A parameter reduction framework was proposed by identifying the most critical heads in each encoder layer using layer-wise relevance propagation and pruning the redundant heads \cite{voita2019analyzing}. It showed that the earlier encoder layer's heads are much more critical than the last. Moreover, in \cite{michel2019sixteen}, it is shown that around 17\% of attention heads can be removed at the test time without significantly impacting inference performance. \textcolor{rev-color}{In \cite{kwon2022fast}, a rapid post-training pruning framework tailored for Transformers is introduced. This framework involves three key fine-tuning stages: mask search, mask rearrangement, and mask tuning. Notably, the method obviates the necessity for any retraining, presenting an efficient approach to achieve model compression in the Transformer architecture. 
}

\subsection{Layer/Block Pruning}
A progressive layer-dropping at the pre-training phase of BERT architecture offers up to 2.5 times speedup at the inference phase with more knowledge transferability and better generalization on downstream tasks \cite{zhang2020accelerating}. FastBERT is proposed with adaptive inference time in which the model uses a sufficient number of encoder layers based on the inference certainty of a sample \cite{liu2020fastbert}. It dynamically adjusts the number of active layers to reduce the overall computational effort. ELBERT proposed the Confidence-Window Based Early Exit mechanism, improving the ALBERT inference speed \cite{xie2021elbert}. In \cite{sajjad2020poor}, it is experimentally shown that using all layers of a pre-trained model in downstream tasks is unnecessary. Studying various layer-dropping strategies shows that many model weights are not necessary. \textcolor{rev-color}{LayerDrop introduced a type of structured dropout that imparts regularization benefits during the training process and enables effective pruning at inference time, resulting in the removal of approximately half of the Transformer layers \cite{fan2019reducing}.}

\subsection{Token/Word-Vector Pruning}
PoWER-BERT proposes a word-vector elimination method that eliminates redundant word-vectors\cite{goyal2020power}. A supplement learnable layer is added between each encoder self-attention and feed-forward layer, and an additional loss function term is used to train these supplement parameters. PoWER-BERT training consists of 3 steps \enquote{fine-tuning, elimination configuration search, and re-training} \cite{goyal2020power}. Funnel-Transformer is inspired by U-net from computer vision \cite{ronneberger2015u}. It improves the Transformer architecture by inserting a pooling layer in each self-attention layer \cite{dai2020funnel}. Funnel-Transformer compresses the sequence of hidden states to a shorter one and reduces the computation cost. \textcolor{rev-color}{Learned Token Pruning (LTP) adaptively removes unimportant tokens as an input sequence passes through transformer layers \cite{kim2022learned}. LTP is applied to RoBERTa \cite{liu2019roberta}, which results in up to 2.0 times throughput improvement with less than 1\% accuracy drop.}

\textcolor{rev-color2}{
In STTABT \cite{lee2022sparse}, the impact of token pruning on later layers' attentions and on final predictions is studied. The study shows that the proposed attention back-tracking allows the model to better retain the full model's performance even at high sparsity rates. Conversely, since in the self-attention architecture all input elements contribute to the formation of each attention output, it can be stated that in the word-vector pruning procedure, no independent concept is discarded. In the subsequent layers, every concept retains the ability to expand and influence the outputs and model predictions. Additionally, STTABT prunes tokens based on a lightweight attention approximation network trained with knowledge distillation, which adds extra training and inference overhead.
}

\textcolor{rev-color2}{
In \cite{anagnostidis2024dynamic}, a learnable mechanism is employed to determine which uninformative tokens can be dropped from the context at any point during the generation process. In \cite{jaegle2021perceiver}, the model leverages an asymmetric attention mechanism to iteratively distill inputs into a tight latent bottleneck, allowing it to scale to handle very large inputs.
}

\subsection{Non-Pruning Methods}
ALBERT is a lighter version of BERT model \cite{lan2019albert}. It uses cross-layer weight sharing and a modified embedding layer structure, making the embedding size independent of the encoder's hidden size. Moreover, the new Sentence Order Prediction (SOP) task is utilized in the pre-training phase instead of the Next Sentence Prediction task. ALBERT reports state-of-the-art results with its parameter reduction and modified pre-training policy. ALBERT is nine times smaller and 1.2 times faster than BERT-base, with less than 2.5\% on average accuracy drop. ALBERT is an example of the parameter-sharing technique.

DistilBERT \cite{sanh2019distilbert}, TinyBERT \cite{jiao2019tinybert} and MobileBERT \cite{sun2020mobilebert} use the knowledge distillation technique during the pre-training and offer up to 5.5 times faster inference on downstream tasks than BERT-base. \textcolor{rev-color}{In \cite{xu2020deep} a Self-Distillation (SD)
mechanism obtains high-accuracy models directly without going through an assistive model.} The main concerns with these techniques are the pre-training effort of the model from scratch, extra training steps, and significant accuracy degradation.

\textcolor{rev-color2}{ 
In \cite{yun2020n}, the authors propose sufficient conditions under which they prove that a sparse attention model can universally approximate any sequence-to-sequence function. They show that sparse Transformers with only $\mathcal{O}_{n}$ connections per attention layer can approximate the same function class as the dense model with $n^2$ connections. Consequently, in recent years, many works \cite{peng2021random, katharopoulos2020transformers, choromanski2020masked, schlag2021linear, wang2020linformer, lee2019set, zaheer2020big, jiang2024minference, zhu2024near} have proposed methods that aim to reduce the complexity of the attention mechanism.
}

\textcolor{rev-color}{
Low-Rank Adaptation (LoRA) involves the freezing of weights in the pre-trained model, and during the fine-tuning phase, it introduces trainable rank decomposition matrices into each layer of the Transformer architecture \cite{hu2021lora}. LoRA enhances the efficiency of memory usage during the training phase by up to 3 times, but it does not confer any advantages during the inference phase. QLoRA improves over LoRA by quantizing the transformer model to 4-bit precision and using paged optimizers to handle memory spikes \cite{dettmers2023qlora}. LQ-LoRA uses an iterative algorithm to decompose each pre-trained matrix into a high-precision low-rank component and a memory-efficient quantized component \cite{guo2023lq}.
}

The previous works are trying to reduce the model size and inference latency. These works show that significant part computations in the models appear unnecessary, and models with less computational complexity can be obtained from the original ones. The main drawbacks of these methods are pre-training from scratch, multi-phase fine-tuning, and significant inference accuracy drop. In these methods, the inference speedup is set in the fine-tuning phase, and the achieved speedup is fixed in the inference phase. The proposed solution tries to mitigate the previous drawbacks. It presents the powerful offline-tuning mechanism that controls the inference speedup of the model after the fine-tuning phase.

\textcolor{rev-color}{
The impact of the speedup methods on the model stability and the input's global context understanding is an important concern that must be studied in solutions evaluations. Backpack learns multiple non-contextual sense vectors for each word in a vocabulary and represents a word in a sequence as a context-dependent, non-negative linear combination of sense vectors in this sequence \cite{hewitt2023backpack}. The Backpack Model shows that each word in the input sequence can interpreted as a sense vector that encodes different aspects of a word which is very sensitive. A good practice for assessing the performance of a method is evaluating it in a more complex task, like text generation tasks that inspect the model proficiency. Therefore the suggested method is applied to the several autoregressive models and evaluated on several text generation tasks.
}

\textcolor{rev-color3}{
Caching methods focus on optimizing the efficiency and speed of inference in autoregressive large language models (LLMs). KV-Runahead enhances prompt phase parallelization by using multiple processes to manage KV-cache population, thereby reducing the Time-to-First-Token (TTFT) \cite{cho2024kv}. CacheGen speeds up context loading with a custom tensor encoder that compresses KV-cache into compact bitstream representations, minimizing decoding overhead \cite{liu2024cachegen}. CacheBlend introduces a method for reusing pre-computed KV-caches and selectively updating a small subset of tokens, optimizing cache reuse \cite{yao2024cacheblend}. Prompt Cache accelerates inference by precomputing and storing attention states for frequently occurring text segments, enabling efficient reuse across different prompts \cite{gim2024prompt}. Lastly, CachedAttention employs a hierarchical KV-caching system using cost-effective memory and storage solutions to manage KV-caches for all requests \cite{gao2024cost}. Together, these methods contribute to faster and more efficient LLM inference by improving cache management and state reuse.
}

\section{Proposed Method}

This section presents details of the proposed method and the analysis and reasoning behind the approach. \textcolor{rev-color3}{Section \ref{background} introduces the preliminary background of the Transformer architecture, while Section \ref{FLOP_Analysis} analyzes the latency and TTFT across various Transformer-based models.} The proposed encoder layer is based on an insight into the amount of the contribution of word-vectors at different layers, which is considered in Section \ref{ACC_Section}. Section \ref{Architecture_Section} explores the proposed architecture in detail. \textcolor{rev-color3}{Additionally, Section \ref{Architecture_Section_Decoder} extends the proposed method to autoregressive architectures and applies it to several newly developed models.}

\subsection{Background}\label{background}

\textcolor{rev-color3}{
Despite the current progress and advancements in mainstream LLM architectures such as GPT, Gemma, Mistral \cite{jiang2023mistral}, and Llama, these models still follow the foundational Transformer architecture \cite{vaswani2017attention}. The primary differences among them lie in the number of parameters, layers, heads, hidden states, and specific modifications to normalization layers, residual paths, and attention implementations. For clarity and a deeper understanding of the suggested method, we first focus on the BERT \cite{lan2019albert} and GPT-2 \cite{radford2019language} models before extending the approach to other large language models (LLMs).
}

\begin{figure}[t]
\centering{\includegraphics[width=0.8\linewidth]{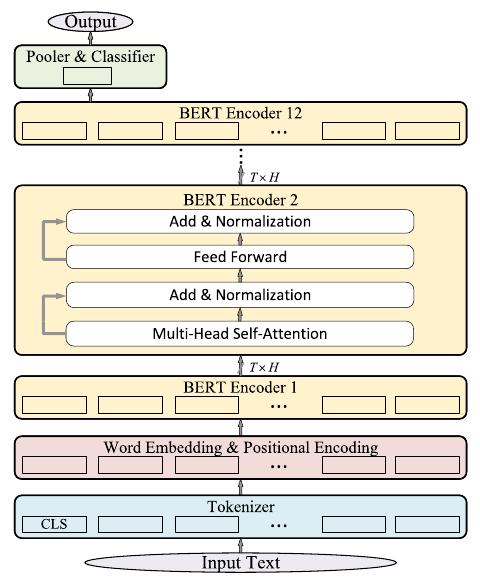}}
\caption{BERT-base \cite{devlin2018bert} architecture with output layers}
\label{fig1}
\end{figure}

Fig.~\ref{fig1} shows BERT-base model architecture and output classifier. The input sentence goes through the tokenizer layer and embedding layer. Then the context is fed into 12 cascaded encoder layers to be processed. The pooler and classifier layers are added in the last layer to perform proper output for the desired task. In the figure, $T$ and $H$ are the numbers of input word-vectors and the hidden state vector size, respectively. Each encoder consists of a \textcolor{rev-color}{Bidirectional} Multi-Head Self-Attention layer and a Feed-Forward layer with several tensor multiplications. The residual branches and normalization layers increase the training stability.

\textcolor{rev-color3}{
\subsection{Transformer Architecture Latency and TTFT Analysis}\label{FLOP_Analysis}
}

\textcolor{rev-color3}{
Table~\ref{tab1} presents an analysis of the latency and TTFT share across the embedding, attention, and feed-forward layers in several Transformer-based models. As shown in the table, regardless of model size, the latency of the embedding layer is minimal, with the majority of latency attributed to the attention and feed-forward layers. Attention share ranges from around 20\% to 47\% of the total. In neural networks, latency and computational effort are closely related to the number of FLOPs. Consequently, reducing the number of FLOPs in the encoder layers, which account for most of the latency, significantly decreases both total latency and TTFT.
}

\begin{table*}[t]
\color{rev-color3}
\caption{Latency and Time-to-First-Token (TTFT) Analysis for BERT \cite{devlin2018bert}, GPT-2 \cite{radford2019language}, T5 \cite{raffel2020exploring}, Gemma2 \cite{team2024gemma2}, \\ Mistral \cite{jiang2023mistral}, and Llama3 \cite{dubey2024llama} Models}
\begin{center}
\renewcommand{\arraystretch}{1.39}
\begin{tabular}{cccccccc}
\hline

\multirow{2}{*}{\textbf{Model}}
    & \multirow{2}{*}{\textbf{Architecture Type}}
    & \multirow{2}{*}{\textbf{Parameters}}
    & \multirow{2}{*}{\textbf{Layers}}
    & \multirow{2}{*}{\textbf{Attention Heads}}
    & \multicolumn{3}{c}{\textbf{Latency/TTFT Share$^{\mathrm{a}}$ (\%)}} \\
    \cline{6-8}

    & & & & & \textbf{Embedding} & \textbf{Attention} & \textbf{Feed-Forward}
    \\ \hhline{========}

    BERT-base & Encoder & 110M & 12 & 12 & 0.58 & 34.94 & 64.48
    \\ \hline
    GPT-2 & Decoder & 124M & 12 & 12 & 0.06 & 34.94 & 65.00
    \\ \hline
    T5-base & Encoder-Decoder & 220M & 12 & 12 & 0.35 & 47.00 & 52.65
    \\ \hline
    T5-large & Encoder-Decoder & 770M & 24 & 16 & 0.06 & 40.46 & 59.47
    \\ \hline
    GPT-2-large & Decoder & 774M & 36 & 20 & 0.01 & 32.25 & 67.74
    \\ \hline
    Gemma2 & Decoder & 2B & 26 & 8 & 0.01 & 20.36 & 79.63
    \\ \hline
    Mistral & Decoder & 7B & 32 & 32 & 0.01 & 20.34 & 79.65
    \\ \hline
    Llama3 & Decoder & 8B & 32 & 32 & 0.01 & 19.55 & 80.44
    \\ \hline

\multicolumn{8}{l}{$^{\mathrm{a}}$ Latency share for BERT model and Time-to-First-Token (TTFT) for other autoregressive models}

\end{tabular}
\label{tab1}
\end{center}
\end{table*}

\subsection{Attention Context Contribution (ACC) metric}\label{ACC_Section}

The context contribution of word-vectors at each self-attention layer is studied in this Section to find more important word-vectors at each encoder layer. For a given encoder layer, the Attention Context Contribution (ACC) metric measures the contribution of word-vectors in the layer's output. This metric is derived from a Score Vector (SV) to find more important word-vectors in each layer. For a given layer, the SV indicates the contribution of each input word-vector to the output of the attention layer.

\begin{figure*}[t]
\centerline{\includegraphics[width=1\linewidth]{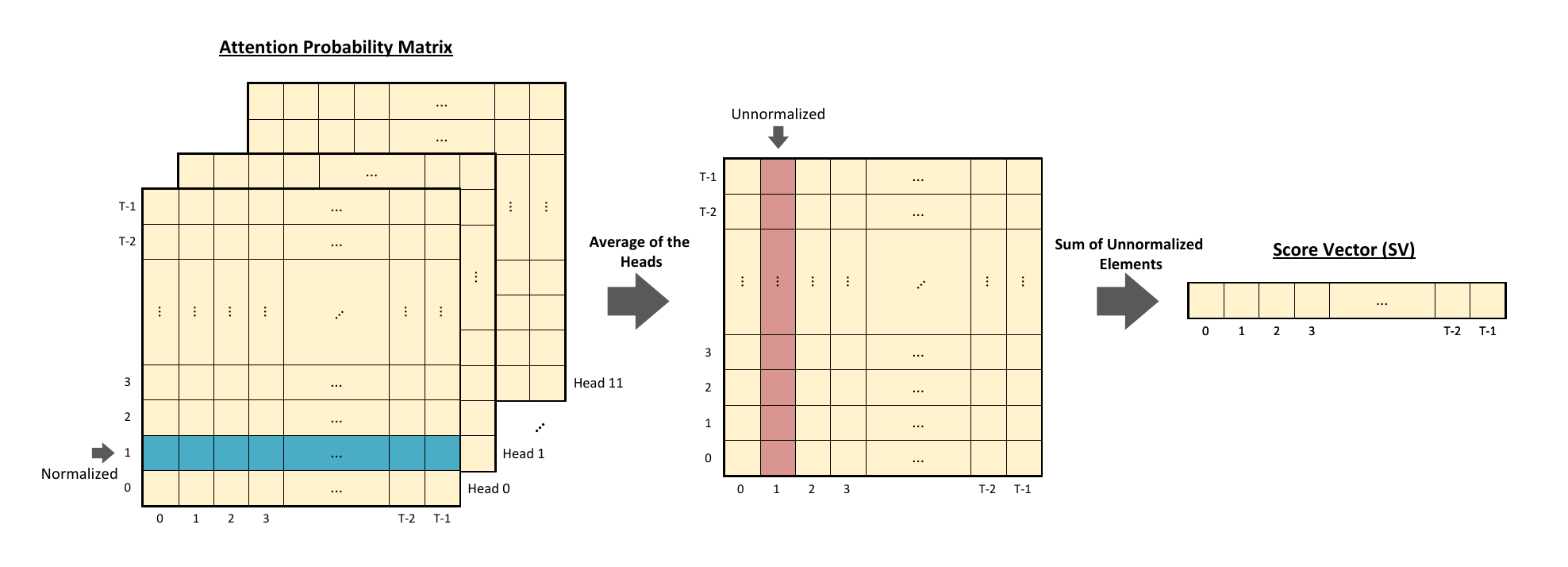}}
\caption{Obtaining Score Vector (SV) from Attention Probability Matrix in a given BERT \cite{devlin2018bert} encoder layer}
\label{fig_ACC}
\end{figure*}

Fig.~\ref{fig_ACC} illustrates the calculation of the Score Vector based on the attention probability matrix. Fig.~\ref{fig_ACC} (left) shows the attention probability matrix of BERT-base model. The rows of this matrix are normalized to one. By averaging the attention probability matrix over the heads, the aggregated probability matrix is resulted, Fig.~\ref{fig_ACC} (center). The SV results from the sum of this matrix along the unnormalized dimension as shown in Fig.~\ref{fig_ACC} (right). The SV elements with higher score values correspond to the word-vectors that contribute more to the layer output.

\begin{figure}[t]
\centerline{\includegraphics[width=0.95\linewidth]{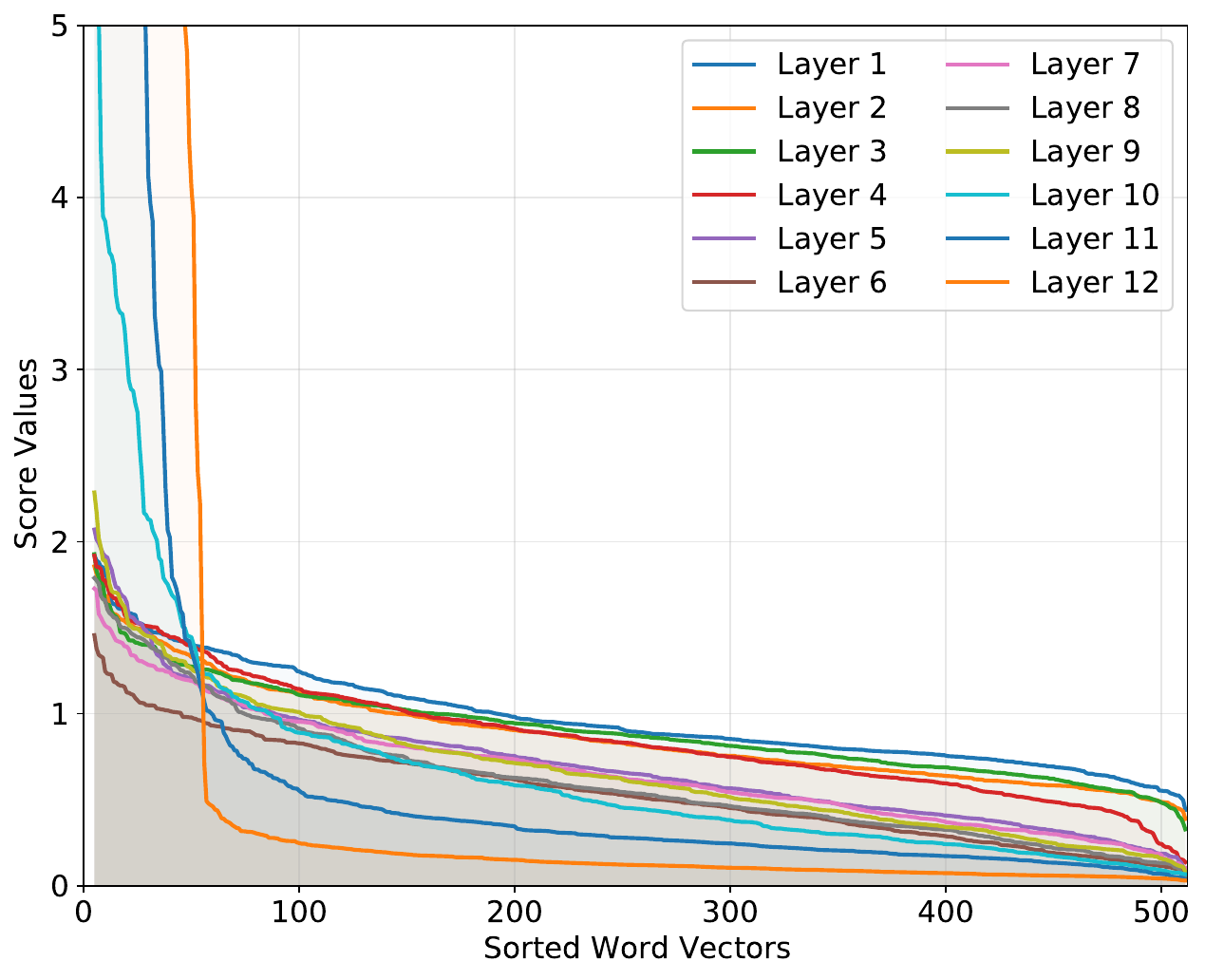}}
\caption{BERT-base\cite{devlin2018bert} Score Vector (SV) results on the IMDB sentiment analysis task\cite{maas2011learning}}
\label{fig_SV}
\end{figure}

Fig.~\ref{fig_SV} illustrates BERT-base SV results on the IMDB sentiment analysis task\cite{maas2011learning} for an input sample. As shown in earlier encoder layers, the SV values approximately have uniform distributions, and in the last encoder layers, the distribution of SV values changes to approximately $\delta$ distribution. The distribution of SV values in the last layers indicates that the concentration of scores is high on a few word-vectors and others have less contribution to the self-attention layer output.

The SV elements are positive and with an average of one. So, the median of SV is bounded. Suppose the median of the SV is lower than its average. In that case, more than half of the elements are lower than the average value, and only a few are higher than the average value. It suggests that most of the word-vectors do not bear enough contribution to the layer output, and a few word-vectors influence the final result of the layer. The median of the Score Vector is interpreted as the proposed Attention Context Contribution metric.

The ACC metric calculation is explained in Algorithm~\ref{alg1}. The algorithm inputs are BERT encoder word-vectors and the model parameters. The algorithm output is the ACC metric vector that suggests the contribution of the word-vectors in each encoder self-attention layer. For encoder layer $l$, the first step is to prepare the pre-computed self-attention probability matrix; $M_{Att}^l$. Then the average of this matrix over its heads is computed; $M_{H}^l$. In the next step, the resulting matrix's summation over its unnormalized dimension is computed; $M_{S}^l$. This matrix is the Score Vector. Attention Context Contribution of the layer ($E_{ACC}^{l}$) is the median of the Score Vector.

\begin{algorithm}[t]
\caption{Attention Context Contribution (ACC) metric}
\begin{algorithmic}[1]
\renewcommand{\algorithmicrequire}{\textbf{Input:}}
\renewcommand{\algorithmicensure}{\textbf{Output:}}
\REQUIRE Encoder inputs $ \in R^{LTH}$  \& fine-tuned BERT \\
\algorithmiccomment{$T$ = number of input word-vectors, $L$ = number of encoder layers, $H$ = hidden state size}
\ENSURE ACC vector: $E_{ACC} \in R^{L}$
\\ \textit{Initialisation}:
\STATE $E_{ACC} = [\emptyset]$
\\ \textit{LOOP Process}
\FOR {$encoder\ l$ in $Encoders$}
\STATE $M_{Att}^l \gets$ Self-Attention Probability Matrix $\in R^{HTT}$
\STATE $M_{H}^l \gets$ Mean of $M_{Att}^l$ over heads $\in R^{TT}$
\STATE $M_{S}^l \gets$ Sum of $M_{H}^l$ over unnormalized dimension $\in R^{T}$
\STATE $E_{ACC}^{l} \gets$ Median of $M_{S}^l$ (Score Vector) $\in R$
\STATE Append $E_{ACC}^{l}$ to $E_{ACC}$
\ENDFOR
\RETURN $E_{ACC}$
\end{algorithmic}
\label{alg1}
\end{algorithm}

\begin{figure*}[t]
\centerline{\includegraphics[width=\linewidth]{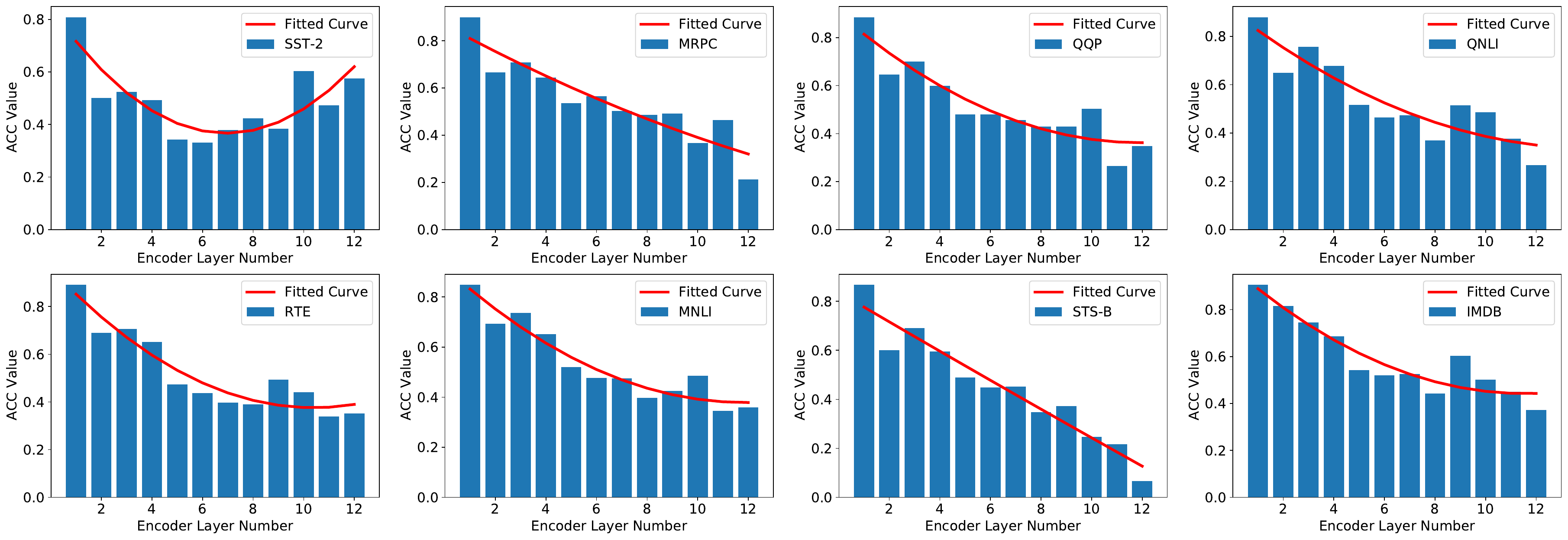}}
\caption{BERT-base \cite{devlin2018bert} Encoder Attention Context Contribution (ACC) metric results (blue bars) and second degree fitted curves (red lines) on SST-2 \cite{socher2013recursive}, MRPC \cite{dolan2005automatically}, QQP \cite{iyer2017first}, QNLI \cite{rajpurkar2016squad}, RTE \cite{bentivogli2009fifth}, MNLI \cite{williams2017broad}, STS-B \cite{cer2017semeval} and IMDB \cite{maas2011learning} tasks.}
\label{fig3}
\end{figure*}

Fig.~\ref{fig3} shows the ACC metric for BERT-base encoder layers on IMDB \cite{maas2011learning} dataset and several General Language Understanding Evaluation (GLUE) \cite{wang2018glue} benchmark tasks. The fitted curve to the ACC metric results shows that the ACC metric is reduced at later layers gradually. It indicates that the fraction of the word-vectors that contribute more in the last layers is less in those layers; hence more word-vectors can be eliminated. By eliminating less important word-vectors in each layer, the processed word-vectors are decreased gradually, and the inference time of the model is decreased.

\textcolor{rev-color}{
As shown in Fig.~\ref{fig3}, the behavior of the ACC metric strongly correlates with the intricate specifications inherent in each task. The ACC results consistently exhibit a monotonic decrease with the layer number, except in the case of the SST-2 \cite{socher2013recursive} dataset, where unexpected behavior is observed. Notably, the SST-2 dataset, centered around sentiment analysis of movie reviews, shares a strong resemblance to the IMDB dataset. A comparative analysis of the ACC curves for SST-2 and IMDB reveals a notable disparity, primarily associated with the length of input sequences. Specifically, in the IMDB dataset, the majority of samples consist of around 512 tokens; conversely, in the SST-2 dataset, the majority of samples feature fewer than 50 tokens. It can be inferred that the limited number of input tokens leads to an undesired sensitivity to word-vectors, and in such cases, some word-vectors are revived through residual branches.
}

\subsection{Proposed Architecture}\label{Architecture_Section}

According to the discussion in Section \ref{FLOP_Analysis}, \textcolor{rev-color3}{attention and feed-forward layers are major contributes of the total latency.} In addition, according to the observation in Section \ref{ACC_Section}, the word-vector contribution in the encoder decreases at later layers. Based on these observations, this Section proposes the details of the proposed architecture.

\begin{figure*}[t]
\centerline{\includegraphics[width=1\linewidth]{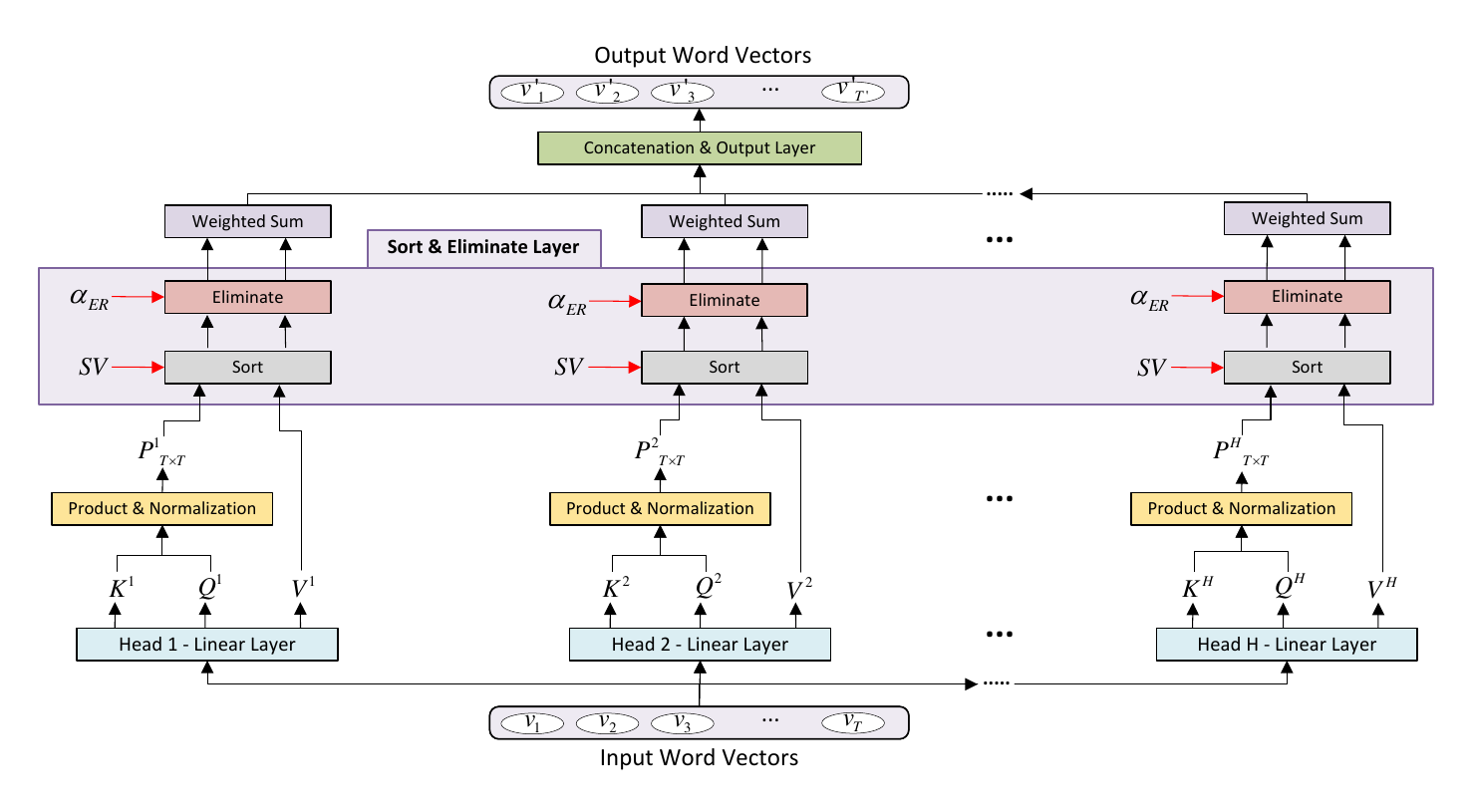}}
\caption{The proposed self-attention layer}
\label{fig_ATT}
\end{figure*}

\textcolor{rev-color3}{In proposed architecture, the orginal BERT self-attention layer is modified with suggested self-attention layer.} Fig.~\ref{fig_ATT} illustrates the proposed multi-head self-attention layer. The proposed architecture adds the Sort and Eliminate layers in the middle of the original BERT multi-head self-attention layer. The Sort layer sorts the input context according to the computed layer Score Vector. The Eliminate layer eliminates a fraction of the sorted context with lower scores according to the Elimination-Rate ($\alpha_{ER}$).

$\alpha_{ER}$ is a hyper-parameter vector that sets the fraction of word-vectors that is eliminated in each encoder layer. For layer $l$, $\alpha_{ER}^l$ is the word-vector Elimination Rate obtained from the pre-computed Elimination-Profile ($\alpha_{EP}$) vector and a pre-specified Speedup-Coefficient hyper-parameter ($\alpha_{SC}$). $\alpha_{SC}$ adjusts the trade-off between accuracy and speed in fine-tuning and offline-tuning phases.

\begin{equation}
\alpha_{ER}^l = \alpha_{EP}^l \times \alpha_{SC}
\label{eq0}
\end{equation}

$\alpha_{EP}^l$ is obtained from the fitted curve to the ACC metric (red lines in Fig.~\ref{fig3}) at layer $l$. In some tasks, the ACC values of layers are not smooth. To increase stability and prevent applying stress to the model in fine-tuning phase, in the proposed formulation, instead of ACC metric values ($E_{ACC}$), the fitted curve to the ACC metric ($P_{ACC}$) is used that presents a more smooth elimination trend. \textcolor{rev-color}{It is important to note that when an increasing behavior of the ACC value occurs in a layer, the elimination process is halted, and the remaining layers proceed using a fixed number of retained word-vectors. In \eqref{eq-1}, the $\alpha_{EP}$ lower bound is set to one to ensure a monotonic descending shape of this hyper-parameter curve. This adaptive approach guarantees the preservation of crucial information and mitigates the impact of unexpected ACC behavior on subsequent layers.}

\begin{equation}
\alpha_{EP}^l = \min{(1,\frac{P_{ACC}^{l}}{P_{ACC}^{l-1}})}
\label{eq-1}
\end{equation}

The number of remaining word-vectors in each encoder layer is calculated according to the number of the layer input word-vectors and the $\alpha_{ER}$ hyper-parameter. Equation \eqref{eq1} presents the number of remaining word-vectors ($T_{l}$) at the output of encoder layer $l$. The minimum number of remaining word-vectors at each encoder output is limited to one.

\begin{equation}
T_{l} = \max{(1, \lfloor \alpha_{ER}^l \times T_{l-1} \rfloor)} \label{eq1}
\end{equation}

The effective number of processed word-vectors can be estimated using the number of word-vectors before and after the eliminate layer. Based on Table~\ref{tab1}, \textcolor{rev-color3}{in BERT-base, around 65\% of latency} occur after the self-attention layer. Equation \eqref{eq2} presents the effective number of Processed word-vectors ($PW_l$) in layer $l$.

\textcolor{rev-color3}{
\begin{equation}
PW_{l} \approx \frac{7 T_{l-1} + 13 T_{l}}{20}
\label{eq2}
\end{equation}
}

Hence,

\textcolor{rev-color3}{
\begin{equation}
PW \approx T 
\left[  \frac{7}{20} + \sum_{i=1}^{L-1}\prod_{j=1}^{i}\alpha_{EP}^{j} + \frac{13}{20}\prod_{j=1}^{L}\alpha_{EP}^{j}  \right]
\label{eq4}
\end{equation}
}

Processed word-vectors in the original BERT-base is:

\begin{equation}
PW_{BERT-base} = T \times L
\label{eq3}
\end{equation}

In these equations, ${L}$ is the number of BERT encoder layers, and ${T}$ is the number of encoder input word-vectors. Since almost all inference latency occurs in the encoder, the overall model computation effort is almost equal to that of the encoder stage. Equation \eqref{eq5} presents the proposed method inference time speedup to BERT-base.

\textcolor{rev-color3}{
\begin{equation}
K_{speedup} \approx \frac{20L}{7 + 20\sum_{i=1}^{L-1}\prod_{j=1}^{i}\alpha_{EP}^{j} + 13\prod_{j=1}^{L}\alpha_{EP}^{j}} 
\label{eq5}
\end{equation}
}

The above equation assumes that the number of remaining word-vectors in each encoder layer is more than a single word-vector. In other words, \eqref{eq1} returns the second argument, which is valid for most cases. For example, if $L$ is 12 and $\alpha_{EP}^{l}$ is constant and equal to 0.8 for all layers, the final inference speed is around \textcolor{rev-color3}{2.99} times faster than BERT-base.

\textcolor{rev-color}{
The proposed Sort and Elimination layers are included in the fine-tuning phase. Given that the eliminated word-vectors constitute a relatively small portion within each layer during the fine-tuning process, we can assume that the gradients effectively propagate through the remaining word vectors. Additionally, the residual paths stay functional, contributing to the updates in the initial layers of the network. This design ensures a robust flow of information throughout the fine-tuning process.
}


To enhance the speed of the suggested self-attention layer, the Sort and Eliminate layers can be positioned after the self-attention concatenation layer. This adjustment results in a reduction of per-head calculations, and for a given layer, the eliminated word-vectors play a role in shaping the layer's output. \textcolor{rev-color}{Furthermore, in accordance with the suggested word-vector elimination strategy, the pruned word-vectors efficiently accelerate the inference phase, leading to an anticipated close alignment between the reduction in Floating Point Operations (FLOPs) and the speedup in inference.}

\textcolor{rev-color3}{
\subsection{Extending Proposed Architecture to Autoregressive Models}\label{Architecture_Section_Decoder}
}

\textcolor{rev-color3}{
The suggested solution can also be applied to decoder architectures, including GPT, Gemma, Mistral, and Llama Large Language Models (LLMs). This section outlines the key considerations for implementing the method on the GPT-2 model, as a representative example of autoregressive models.
}

\textcolor{rev-color}{
The foundational GPT-2 model comprises 12 decoder layers, with each decoder featuring a causal attention layer equipped with 12 heads. The total number of parameters in the GPT-2 model is approximately \textcolor{rev-color3}{124 million}. A key distinction between the architectures of GPT-2 and the BERT-base model lies in the design of their attention layers. As explained in Section \ref{background}, the attention layers of BERT-base encoders are bidirectional, involving all input word-vectors in generating each output word-vector. Conversely, the GPT-2 model, tailored for text generation tasks, employs causal attention layers, where only the preceding word-vectors contribute to the attention layer output at each time stamp.
}

\textcolor{rev-color}{
In a causal attention layer, the attention probability matrix, represented as $M_{Att}^l$, adopts the form of a triangular matrix. To extend the proposed architecture outlined in Section \ref{Architecture_Section} to causal attention layers, an additional normalization step needs to be integrated after the fifth step in Algorithm~\ref{alg1}. This supplementary step normalizes the Score Vector $M_{S}^l$ based on the count of non-zero elements in the probability matrix $M_{Att}^l$. Equation \eqref{eq6} articulates this normalization process in layer $l$, wherein the $i$-th element of the Score Vector is adjusted relative to the number of non-zero elements in the $i$-th column of the probability matrix. Through this normalization step, the resulting score values become comparable.
\begin{equation}
M_{S_i}^{l'} = M_{S_i}^l \div N_{M Att_i}^l
\label{eq6}
\end{equation}
}

\section{Experiments}\label{EXP_Section}

\textcolor{rev-color3}{
In this section, we implement and evaluate the suggested method on BERT-base, GPT-2, Flan-T5 \cite{chung2024scaling}, Gemma2 \cite{team2024gemma2}, Mistral, and Llama3 \cite{dubey2024llama}, exploring its effectiveness across various natural language understanding tasks, text generation datasets, and instruction-tuning benchmarks. Toward the conclusion of this section, we examine the offline-tuning property and conduct an ablation study on the proposed attention layer.
}

\subsection{Datasets}

\textcolor{rev-color3}{The performance of the proposed method on natural language understanding tasks} is evaluated using the IMDB dataset \cite{maas2011learning}, which focuses on sentiment analysis, as well as the General Language Understanding Evaluation (GLUE) benchmark \cite{wang2018glue}. The GLUE benchmark includes two single-sentence classification tasks: SST-2 \cite{socher2013recursive} and CoLA \cite{warstadt2019neural}; two multi-sentence similarity classification tasks: MRPC \cite{dolan2005automatically} and QQP \cite{iyer2017first}; one binary question answering task: QNLI \cite{rajpurkar2016squad}; three natural language inference tasks: RTE \cite{bentivogli2009fifth}, MNLI-M, and MNLI-MM \cite{williams2017broad}; and one multi-sentence similarity regression task: STS-B \cite{cer2017semeval}.

\textcolor{rev-color3}{To assess the performance of the approach on text generation tasks}, we conduct evaluations on several datasets. The WikiText-103 \cite{merity2016pointer} language modeling dataset, comprising over 100 million tokens extracted from verified Good and Featured articles on Wikipedia, serves as a primary evaluation ground. Additionally, the PTB \cite{marcus1993building} dataset, featuring a million words from 1989 Wall Street Journal material, and the 1BW \cite{chelba2014billion} dataset, established as a benchmark corpus for measuring progress in statistical language modeling, are included in our evaluation. Moreover, we assess the approach on The LAMBADA \cite{paperno-EtAl:2016:P16-1} dataset, a collection of narrative passages characterized by the intriguing feature that human subjects can predict their last word when exposed to the entire passage. Extracted from BookCorpus, The LAMBADA dataset comprises 10,022 passages, contributing to a comprehensive evaluation of the proposed method's performance in decoder models.

\textcolor{rev-color3}{
To further demonstrate the suggested approach's performance in language understanding, it is evaluated on instruction-tuning tasks and assessed using the Massive Multitask Language Understanding (MMLU) benchmark. This benchmark includes exam questions from 57 diverse tasks, covering subjects such as mathematics, history, law, and medicine \cite{hendrycks2020measuring}.
}

\subsection{Models Settings}\label{Settings_Section}

\textcolor{rev-color}{In the assessment of the method on BERT-base, the encoder self-attention layer undergoes substitution with the proposed encoder self-attention layer.} 
The hyper-parameters are the same as those in BERT-base with extra $\alpha_{EP}$, and $\alpha_{ER}$ hyper-parameters are added to set the word-vector elimination rate in each encoder layer. The hyper-parameters are: learning rate - [\num{1.5e-5}, \num{4.5e-5}], batch size - \{4, 8, 16, 32, 64, 128\}, number of epochs - \{2, 3, 4, 5\}, Dropout - [0.1, 0.15], $\alpha_{ER}$ - [0.77, 0.97], and $\alpha_{EP}$ vector that is obtained from \eqref{eq-1}. The cross-entropy loss function is used for classification tasks and the Mean Square Error (MSE) loss function for regression tasks. The linear learning rate scheduler is used for all tasks. Initial linear learning warm-up is used for some tasks. In all tasks, AdamW optimizer with $\epsilon=\num{1e-8}$, $\beta1=0.9$ and $\beta2=0.999$ is used. The model input sequence length is delicately fixed so that less than 1\% of samples are truncated. The selected input sentence length for each dataset is specified in Table~\ref{tab2}.

\textcolor{rev-color}{
In the case of the GPT-2 model, the attention layer within each decoder layer undergoes modification with the proposed attention layer, which is described in Section \ref{Architecture_Section_Decoder}. Introducing the hyper-parameters $\alpha_{EP}$ and $\alpha_{ER}$ becomes essential to determine the word-vector elimination rate in each encoder layer. The specified hyper-parameters for the evaluation are as follows: a learning rate of \num{1e-6}, a batch size of 8, a three-epoch training duration, a dropout rate of 0.1, $\alpha_{ER}$ values ranging between 0.9 and 1.2, and an $\alpha_{EP}$ vector derived from equation \eqref{eq-1}. These parameters collectively define the experimental setup for a comprehensive evaluation of the method's impact on the GPT-2 model.
}

\textcolor{rev-color3}{
For the instruction-tuning evaluation, the Flan-T5\footnote{Flan-T5 is an enhanced version of T5 \cite{raffel2020exploring}}, Gemma2, Mistral, and Llama3 models are targeted.} In these models, the self-attention layers are replaced with the proposed self-attention layer, and other layers are adjusted to accommodate a dynamic number of word vectors in each encoder and decoder layer. The models use their default weights without any tuning or fine-tuning. The parameter $\alpha_{ER}$ is selected within the range [0.8, 1.0], and all other evaluation hyperparameters remain consistent with those of the default models.

\subsection{Implementation Details}

\textcolor{rev-color}{
The initial models utilized for this study are the pre-trained versions of BERT-base, GPT-2, Flan-T5, \textcolor{rev-color3}{Gemma2, Mistral, and Llama3} from the Hugging Face\footnote{https://huggingface.co/} implementations. The implementation is carried out in the PyTorch\footnote{https://pytorch.org/} library, utilizing an Nvidia Tesla P100 GPU for BERT-base, an Nvidia Tesla T4 GPU for GPT-2, \textcolor{rev-color3}{and an Nvidia Tesla A100 GPU for Flan-T5, Gemma2, Mistral, and Lllama}. The datasets are sourced from the Hugging Face Datasets library, and the GLUE results are validated through the official GLUE service\footnote{https://gluebenchmark.com/}. \textcolor{rev-color2}{The instruction tuning assessment is done with help of the INSTRUCTEVAL \cite{chia2023instructeval} method.} This framework ensures consistency and reliability in the assessment of the proposed solution's performance on the specified models.}

\textcolor{rev-color3}{\subsection{Experimental Results on Language Understanding Tasks}}

The experimental results of \textcolor{rev-color3}{the proposed method applied to BERT-base} for a wide range of tasks are presented in Table~\ref{tab2}. The proposed approach is 2.9 times faster than BERT-base, with only a 0.7\% accuracy drop on the test set. Additionally, the model's inference time is up to 4.8 times faster than BERT-base, with just a 0.9\% reduction in accuracy. These results demonstrate that the suggested method significantly enhances the performance and resource efficiency of Transformer-based models while maintaining minimal accuracy loss.

The table shows the experimental speedup, and the expected speedup. \textcolor{rev-color3}{The inference speedup is consistent with expected speedup predicted using \eqref{eq5}.} The slight differences are due to the computational overheads, including the computational effort of the tokenizer and output classifier layers that are not considered in \eqref{eq5}. The consistency between the experimental speedup, the expected speedup, approves the hypothesis discussed in Section \ref{Architecture_Section}.

\begin{table*}[t]
\color{rev-color3}
\caption{Experimental Results of the Proposed Method on the GLUE \cite{wang2018glue} Benchmark and IMDB \cite{maas2011learning} \\ Dataset with BERT-base as the Target Model}
\begin{center}
\renewcommand{\arraystretch}{1.39}
\begin{tabular}{cccccccccccc}
\hline
\textbf{Dataset$^{\mathrm{a}}$} 
    & \textbf{CoLA}
    & \textbf{SST-2}
    & \textbf{MRPC}
    & \textbf{QQP}
    & \textbf{QNLI}
    & \textbf{RTE}
    & \textbf{MNLI\textsubscript{M}}
    & \textbf{MNLI\textsubscript{MM}}
    & \textbf{STS-B}
    & \textbf{IMDB}
    & \textbf{Avg}
    \\ \hhline{============}

BERT-base & 52.7 & 93.5 & 88.7 & 71.2 & 91.0 & 68.9 & 84.7 & 83.8 & 84.3 & 93.7 & 81.25 \\
This Work & 53.4 & 92.6 & 88.0 & 70.1 & 90.2 & 67.9 & 83.7 & 82.7 & 84.2 & 92.7 & 80.55 \\ \hline
Expected Speedup$^{\mathrm{b}}$ & 3.7x & 2.4x & 2.9x & 4.8x & 2.3x & 2.6x & 2.5x & 2.5x & 2.2x & 2.6x & 2.8x \\
Inference Speedup & 3.8x & 2.4x & 2.9x & 4.7x & 2.3x & 2.6x & 2.5x & 2.5x & 2.3x & 2.7x & 2.9x \\ \hline
Input Seq Length & 64 & 64 & 128 & 128 & 128 & 256 & 128 & 128 & 64 & 512 & 160 \\ \hline

\multicolumn{12}{p{0.88\linewidth}}{$^{\mathrm{a}}$ Matthew's Correlation is reported for CoLA, F1-score for MRPC and QQP, Pearson Correlation for STS-B, and Accuracy for the remaining tasks.}\\
\multicolumn{12}{l}{$^{\mathrm{b}}$ \thead[l]{Using equation \eqref{eq5}.}}
\end{tabular}
\label{tab2}
\end{center}
\end{table*}

Table~\ref{tab3} compares the proposed approach and the previously reported \textcolor{rev-color3}{speedup} methods. BERT-base model ($L$=12, $h$=12, $d$=768) is used as the basis for comparison. The test results, the average accuracy drop compared to BERT-base, the average number of FLOPs, and the average inference speedup are reported for each model. As shown in the table, the proposed approach achieves 2.9 times speedup with 0.73\% on average accuracy drops compared to BERT-base model.

\begin{table*}[t]
\color{rev-color3}
\caption{
Comparison of Results Between BERT\textsubscript{base} \cite{devlin2018bert}, DistilBERT \cite{sanh2019distilbert}, TinyBERT \cite{jiao2019tinybert}, MobileBERT \cite{sun2020mobilebert}, PoWER-BERT \cite{goyal2020power}, \\ LayerDrop \cite{fan2019reducing}, FPT \cite{kwon2022fast}, and Latency-Adjustable BERT\textsubscript{base} (This Work)}
\begin{center}
\renewcommand{\arraystretch}{1.39}
\begin{tabular}{ccccccccccccc}
\hline
\textbf{Model$^{\mathrm{a}}$} 
    & \textbf{Method}
    & \textbf{CoLA}
    & \textbf{SST-2}
    & \textbf{MRPC}
    & \textbf{QQP}
    & \textbf{QNLI}
    & \textbf{RTE}
    & \textbf{MNLI\textsubscript{M}}
    & \textbf{MNLI\textsubscript{MM}}
    & \textbf{Avg Drop}
    & \textbf{FLOPs}
    & \textbf{Speedup}
    \\ \hhline{=============}
    
BERT-base & - & 52.7 & 93.5 & 88.7 & 71.2 & 91.0 & 68.9 & 84.7 & 83.8 & 0.00 & 22.5B & 1.0x \\ \hline
DistilBERT & KD & 51.3 & 91.3 & 87.5 & 70.1 & 89.2 & 59.9 & 82.2 & - & 2.74 & 11.3B & 1.7x \\ \hline
TinyBERT & KD & 51.1 & 93.1 & 87.3 & 71.6 & 90.4 & 70 & 84.6 & 83.2 & 0.40 & 11.3B & 2.0x \\ \hline
MobileBERT$^{\mathrm{b}}$ & KD & 50.5 & 92.8 & 88.8 & 70.2 & 90.6 & 66.2 & 83.3 & 82.6 & 1.19 & 5.7B & 5.5x \\ \hline
PoWER-BERT$^{\mathrm{c}}$ & Pruning & 52.3 & 92.1 & 88.1 & 70.2 & 90.1 & 67.4 & 83.8 & 83.1 & 0.92 & - & 2.9x \\ \hline
LayerDrop & Pruning & - & 92.5 & - & - & 89.4 & - & 82.9 & - & 1.47 & 11.3 & 1.7x \\ \hline
FPT & Pruning & - & 92.6 & - & - & 90.3 & - & 83.5 & - & 0.93 & 14.5 & 1.5x \\ \hline
This Work & Pruning & 53.4 & 92.6 & 88.0 & 70.1 & 90.2 & 67.9 & 83.7 & 82.7 & \textbf{0.73} & \textbf{7.7B} & \textbf{2.9x} \\ \hline

\multicolumn{13}{l}{$^{\mathrm{a}}$ Matthew's Correlation is reported for CoLA, F1-score for MRPC and QQP, Pearson Correlation for STS-B, and Accuracy for the remaining tasks.}\\ 
\multicolumn{13}{l}{$^{\mathrm{b}}$ MobileBERT starts from the BERT-large architecture, which has approximately three times as many parameters as BERT-base.} \\
\multicolumn{13}{l}{$^{\mathrm{c}}$ PoWER-BERT training involves three fine-tuning steps, and the model's inference speedup is fixed during inference.}
\end{tabular}
\label{tab3}
\end{center}
\end{table*}

Compared to other works, DistilBERT model achieves 1.7 times speedup with around 2.74\% accuracy degradation on average. In DistilBERT, a pre-trained general-purpose language presentation model must be trained from scratch with significant effort. The TinyBERT model speedup is 2.0 times with 0.4\% accuracy drops on average. TinyBERT uses a two-stage learning framework that performs Transformer distillation at pre-training and fine-tuning phases with considerable excessive computational effort. The MobileBERT model speedup is 5.5 times with around 1.19\% accuracy drop, but the starting point of this model is BERT\textsubscript{large}. The PoWER-BERT model speedup is 2.9 times with around 0.92\% accuracy drop. However, the training of PoWER-BERT consists of 3 phases: "fine-tuning, elimination configuration search, and re-training" \cite{goyal2020power} that impose extra training effort. \textcolor{rev-color}{LayerDrop and FPT speedups are 1.7 and 1.5 times, respectively, with corresponding average accuracy drops of 1.47\% and 0.93\%.}

The table shows that the results of the suggested technique are competitive and more effective than previous baseline methods across a wide range of experimental tasks. Furthermore, compared to recently reported methods, our approach is applied only during the fine-tuning phase and offers the offline-tuning property. \textcolor{rev-color2}{Blank fields in the table, indicate instances where some works did not report their assessments on certain datasets.}

To assess the impact of the proposed method on GPT-2 model, the training process involves utilizing 1\% of the OpenWebText \cite{Gokaslan2019OpenWeb} dataset with a stride of 128 for three epochs. Table~\ref{tab6} exhibits the classification accuracy findings of the proposed method in contrast to GPT-2 and DistilGPT2. Notably, the proposed method attains an inference latency three times quicker than baseline model, accompanied by a marginal average accuracy reduction of 0.55\%. Blank fields in the tables, indicate instances where some works did not report their assessments on certain datasets.

\begin{table}[t]
\color{rev-color3}
\caption{Accuracy comparison between GPT-2 \cite{radford2019language} and DistilGPT2 \cite{sanh2019distilbert} with latency adjustable GPT-2 (This Work)}
\begin{center}
\renewcommand{\arraystretch}{1.39}
\begin{tabular}{ccccc}
\hline
\textbf{Model} 
    & \textbf{\thead{SST-2}}
    & \textbf{\thead{IMDB}}
    & \textbf{\thead{Avg Drop}}
    & \textbf{\thead{Speedup}}
    \\ \hhline{=====}

GPT-2 & 93.2 & 94.1 & 0.00 & 1.0x \\ \hline
DistilGPT2 & 91.5 & 92.1 & 1.85 & 1.9x \\ \hline
This Work & \textbf{92.8} & \textbf{93.4} & \textbf{0.55} & \textbf{3.0x} \\ \hline

\end{tabular}
\label{tab6}
\end{center}
\end{table}

\textcolor{rev-color3}{\subsection{Experimental Results on Text Generation Tasks}}

\textcolor{rev-color}{
Table~\ref{tab5} presents the results of the zero-shot perplexity assessment for the suggested method in text generation tasks, offering a basis for comparison with other contemporary studies. The findings demonstrate that the method effectively maintains the global context of the input sequence. The proposed Attention Context Contribution (ACC) metric performs well, ensuring that the elimination of word-vectors does not compromise the input's global context. It is noteworthy that the original GPT-2 model, trained by OpenAI, benefited from more extensive hardware and dataset resources. However, our proposed architecture enables efficient adaptation with just a few small epochs. Furthermore, the proposed technique exhibits superior performance when compared to the DistilGPT2 \cite{sanh2019distilbert} and ZipGPT2 \cite{kurtic2023ziplm} models in the evaluation.
}

\begin{table}[t]
\color{rev-color3}
\caption{
Perplexity (PPL) Comparison Between GPT-2 \cite{radford2019language}, DistilGPT2 \cite{sanh2019distilbert}, ZipGPT2 \cite{kurtic2023ziplm}, and Latency-Adjustable GPT-2 \\ (This Work)}

\begin{center}
\renewcommand{\arraystretch}{1.39}
\begin{tabular}{cccccc}
\hline
\textbf{Model} 
    & \textbf{\thead{Wiki\\Text-103$^{\mathrm{a}}$}}
    & \textbf{\thead{LAMBADA}}
    & \textbf{\thead{PTB}}
    & \textbf{\thead{1BW$^{\mathrm{b}}$}}
    & \textbf{Speedup}
    \\ \hhline{======}

GPT-2 & 37.5 & 35.1 & 36.46 & 45.54 & 1.0x \\ \hline
DistilGPT2 & 43.0 & 70.43 & 63.57 & 75.40 & 1.9x \\ \hline
ZipGPT2 & 72.1 & - & - & - & 3.3x \\ \hline

\multirow{2}{*}{This Work}  & \textbf{43.2} & \textbf{59.0} & \textbf{52.9} & \textbf{51.66} & \textbf{2.7x} \\
                            & \textbf{49.3} & \textbf{64.6} & \textbf{61.7} & \textbf{54.53} & \textbf{3.5x} \\ \hline

\multicolumn{6}{l}{$^{\mathrm{a}}$ Lower perplexity is better.} \\ 
\multicolumn{6}{l}{$^{\mathrm{b}}$ The models are evaluated on 0.5\% of the 1BW test set.}

\end{tabular}
\label{tab5}
\end{center}
\end{table}

\textcolor{rev-color3}{\subsection{Experimental Results on Instruction Tuning Tasks}}

\textcolor{rev-color3}{
For the instruction-tuning evaluation, the Flan-T5, Gemma2, Mistral, and Llama3 models are selected. In these models, the self-attention layers are replaced with the proposed encoder self-attention layer, and the default weights are used without any additional tuning or fine-tuning. The method's performance is evaluated using various prompting types on the MMLU benchmark. Table~\ref{tab7} presents the experimental results for 5-shot direct prompting on the MMLU benchmark. As shown, the suggested method achieves up to a 3-fold reduction in TTFT with only a marginal decrease in score. 
}

\begin{table}[t]
\color{rev-color3}
\caption{Time-to-First-Token (TTFT) Speedup Results of the Proposed Approach on Flan-T5 \cite{chung2024scaling}, Gemma2 \cite{team2024gemma2}, Mistral \cite{jiang2023mistral}, and Llama3 \cite{dubey2024llama} Evaluated with the MMLU \cite{hendrycks2020measuring} \\ Instruction Tuning Dataset.}

\begin{center}
\renewcommand{\arraystretch}{1.39}
\begin{tabular}{ccccc}
\hline
\textbf{\thead{Architecture \\ Name}} 
    & \textbf{Model}
    & \textbf{\thead{Model \\ Size}}
    & \textbf{\thead{MMLU \\ Score$^{\mathrm{a}}$}}
    & \textbf{\thead{TTFT \\ Speedup$^{\mathrm{b}}$}}
    \\ \hhline{=====}

\multirow{4}{*}{\thead{T5}}
    & \multirow{2}{*}{Flan-T5-base} & \multirow{2}{*}{220M} & 34.05 & 1.00x \\
    & & & 32.24 & 1.54x \\
    & \multirow{2}{*}{Flan-T5-large} & \multirow{2}{*}{770M} & 41.95 & 1.00x \\
    & & & 41.51 & 1.95x \\ \hline
\multirow{2}{*}{\thead{Gemma}}
    & \multirow{2}{*}{Gemma2} & \multirow{2}{*}{2B} & 56.74 & 1.00x \\
    & & & 56.20 & 1.63x \\ \hline
\multirow{2}{*}{\thead{Mistral}}
    & \multirow{2}{*}{Mistral} & \multirow{2}{*}{7B} & 41.85 & 1.00x \\
    & & & 40.83 & 1.65x \\ \hline
\multirow{2}{*}{\thead{Llama}}
    & \multirow{2}{*}{Llama3} & \multirow{2}{*}{8B} & 65.69 & 1.00x \\
    & & & 64.91 & 2.86x \\ \hline

\multicolumn{5}{p{0.9\linewidth}}{$^{\mathrm{a}}$ MMLU includes exam questions from 57 tasks, uses 5-shot direct prompting, and measure the exact-match score.}\\
\multicolumn{5}{p{0.9\linewidth}}{$^{\mathrm{b}}$ The proposed attention layer is applied to the models, and the Time-to-First-Token (TTFT) speedup is reported.}
\end{tabular}
\label{tab7}
\end{center}
\end{table}

\subsection{Offline-Tuning Property}

\begin{figure*}[t]
\centering
\subfloat[]{\includegraphics[width=0.3\linewidth]{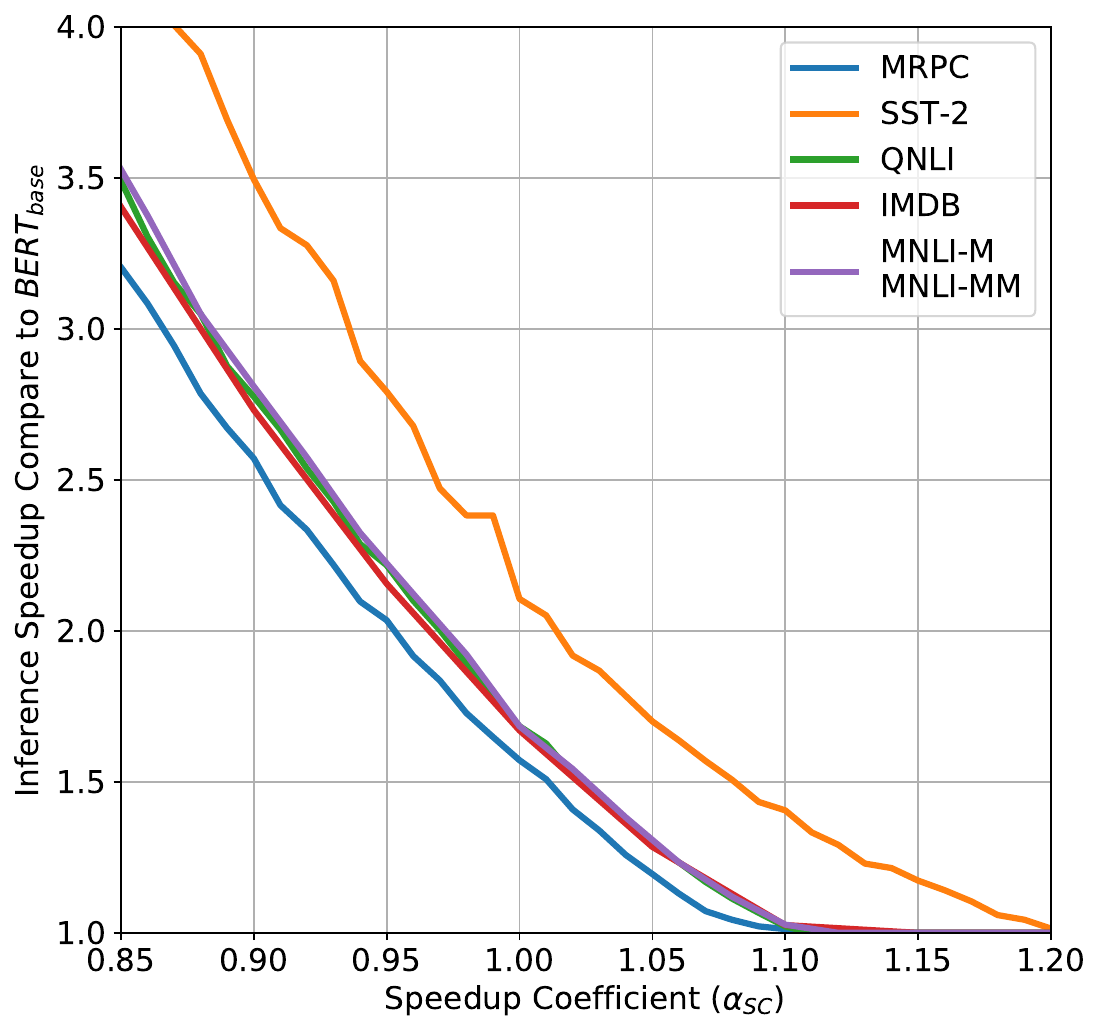}%
\label{offline_stb_1}}
\hfil
\subfloat[]{\includegraphics[width=0.3\linewidth]{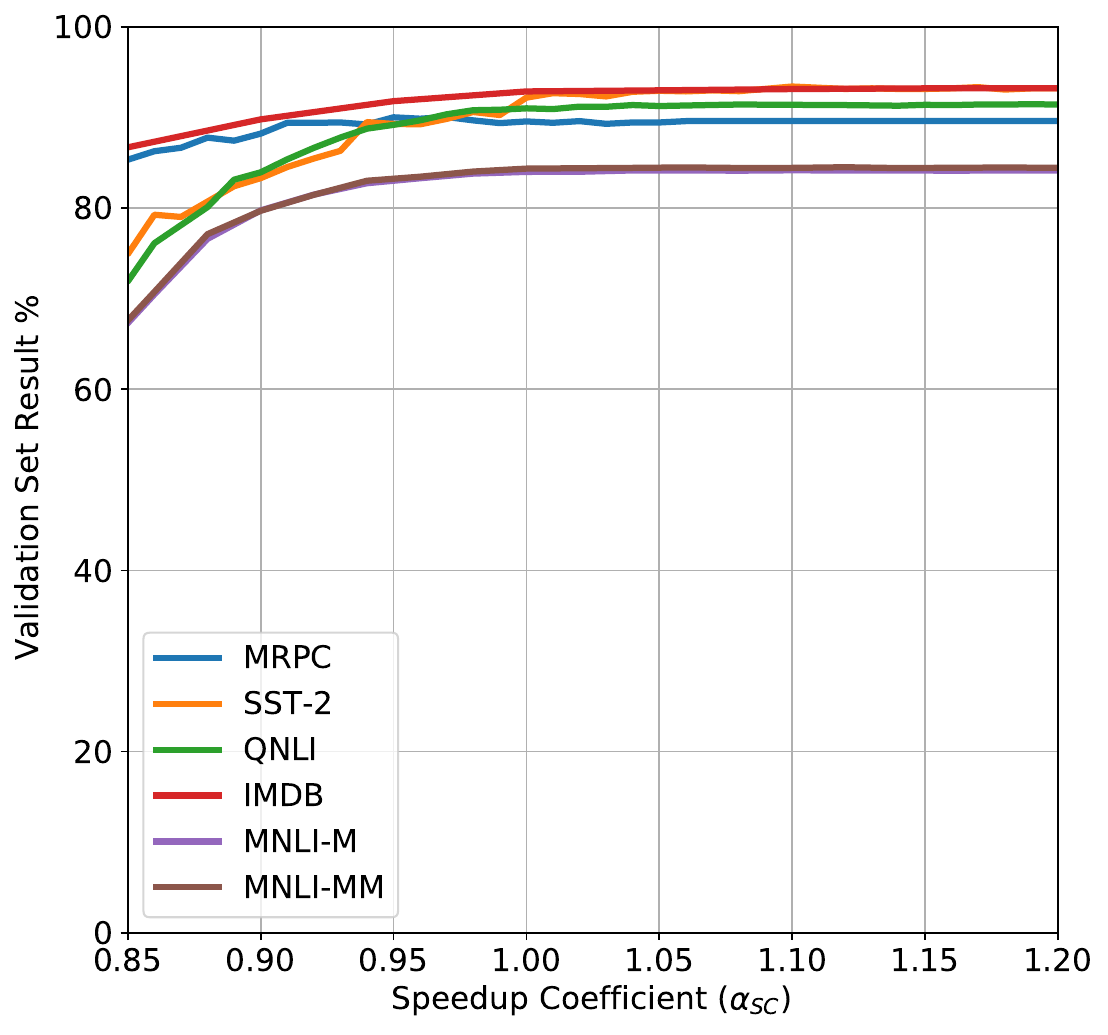}%
\label{offline_stb_2}}
\hfil
\subfloat[]{\includegraphics[width=0.3\linewidth]{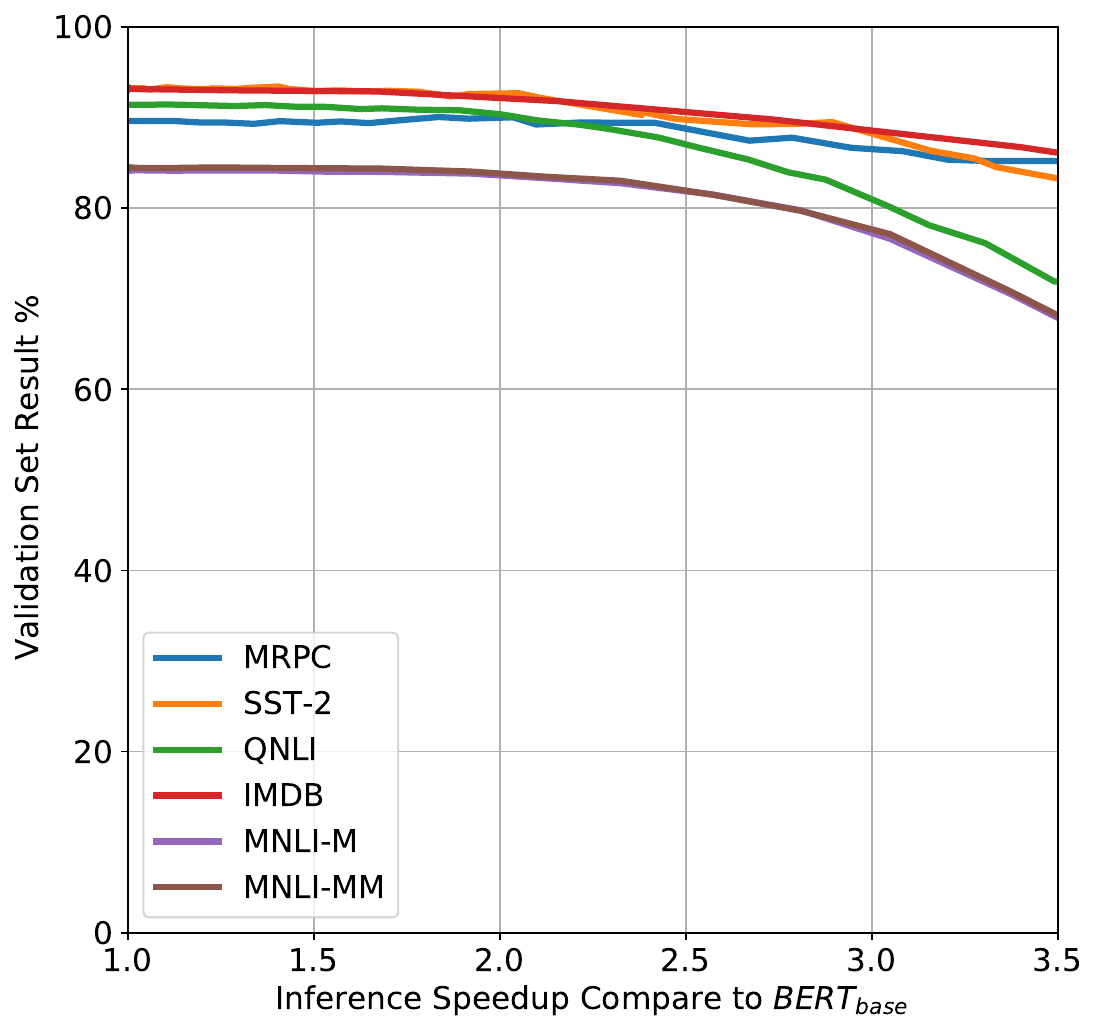}%
\label{offline_stb_3}}
\caption{The proposed offline-tuning results, (a)Inference speedup compare to BERT-base versus $\alpha_{SC}$, (b)Validation set results versus $\alpha_{SC}$, (c)Validation set results versus speedup compare to BERT-base for MRPC \cite{dolan2005automatically}, SST-2 \cite{socher2013recursive}, QNLI \cite{rajpurkar2016squad}, IMDB \cite{maas2011learning} and MNLI-M/MM \cite{williams2017broad} tasks.}
\label{fig5}
\end{figure*}

After the fine-tuning phase, the trade-off between inference accuracy and speed of the model can be adjusted with the offline-tuning property. This property controls each encoder layer's word-vector elimination rate by adjusting the Speedup-Coefficient hyper-parameter ($\alpha_{SC}$) value. Fig.~\ref{fig5} presents the offline-tuning property results on BERT-base. In the following experiments, the model first fine-tuned with $\alpha_{SC}$ equal to one for a given task, and then in the offline-tuning phase, the $\alpha_{SC}$ value changes from 0.85 to 1.2 to observe its effect on inference speedup and accuracy.

Fig.~\ref{offline_stb_1} shows the model inference speedup for a range of $\alpha_{SC}$ hyper-parameter values on multiple tasks. The model inference speedup increased more than three times compared to BERT-base, with the $\alpha_{SC}$ value decreasing to 0.85. Fig.~\ref{offline_stb_2} presents the accuracy stability on multiple tasks. It shows that the inference results are kept stable over a wide range of $\alpha_{SC}$ values selections. Fig.~\ref{offline_stb_3} shows that the inference results are almost unchanged even if the complexity of the model reduces by 70\% according to the proposed method. In these experiments, the model's accuracy on validation sets is reported because of the GLUE test set verification limitations.

Two approaches can be offered to select the proper value of the $\alpha_{SC}$ hyper-parameter in the fine-tuning phase. The first approach is using equation \eqref{eq5} and calculating the approximate value of the $\alpha_{SC}$ based on the desired inference speedup. The second approach is to select the $\alpha_{SC}$ value using suggested Fig.~\ref{fig5}(c) curves. After the fine-tuning phase, the speedup can be accurately adjusted with the offline-tuning property.

\begin{figure}[t]
\color{rev-color}
\centerline{\includegraphics[width=0.85\linewidth]{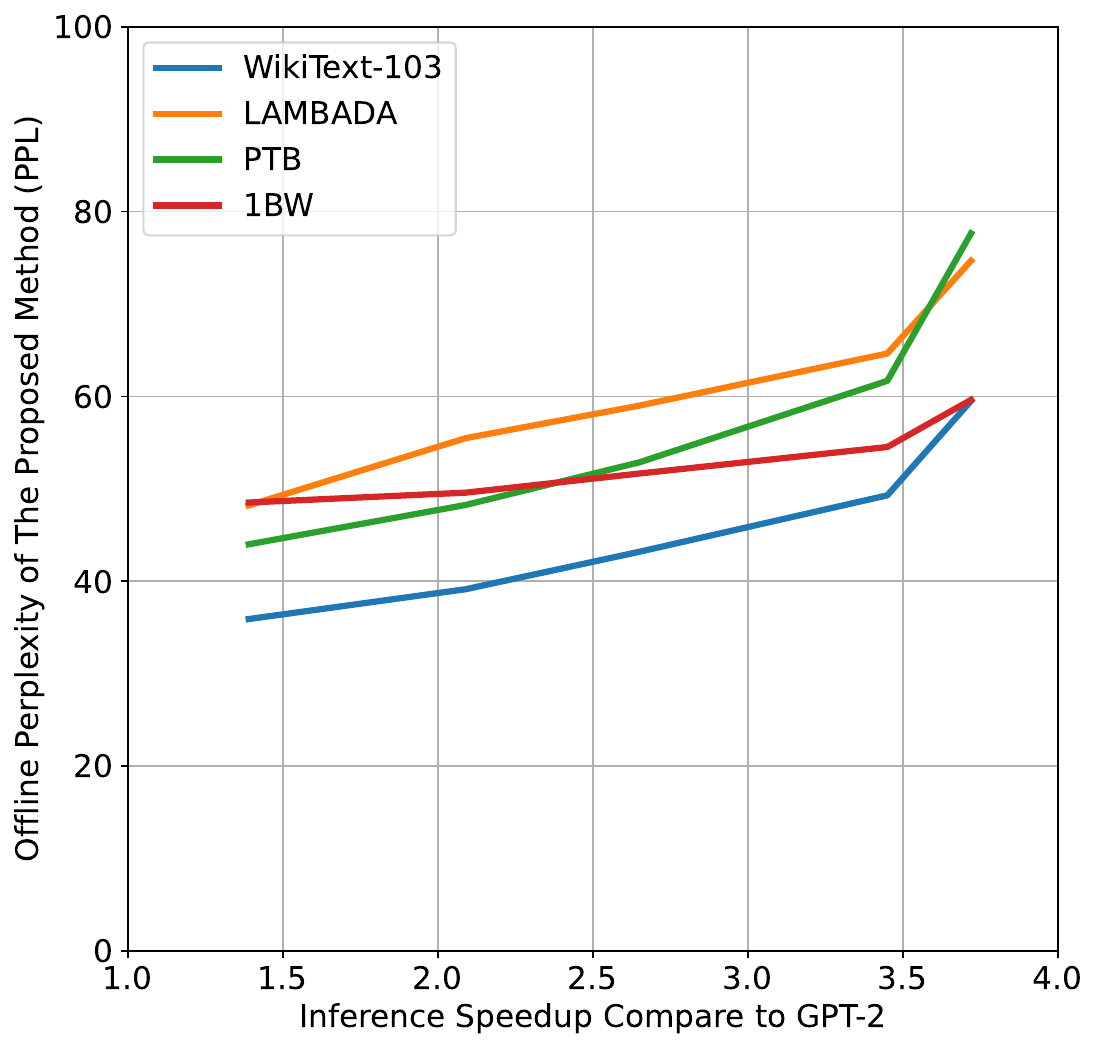}}
\caption{Offline Perplexity (PPL) of the proposed method on WikiText-103 \cite{merity2016pointer}, LAMBADA \cite{paperno-EtAl:2016:P16-1}, PTB \cite{marcus1993building} and 1BW \cite{chelba2014billion} datasets.}
\label{fig_STB_PPL}
\end{figure}

\textcolor{rev-color}{
The offline perplexity (PPL) stability of the proposed method for text generation tasks is evaluated by training the model with a specific speedup coefficient and then modifying this value during inference, while measuring the model's zero-shot perplexity. Fig.~\ref{fig_STB_PPL} illustrates the method's offline perplexity results on GPT-2 across various datasets, including WikiText-103, LAMBADA, PTB, and 1BW. The graph demonstrates the model's robust stability, showing a smooth perplexity curve across a wide range of speedup values, which highlights the solution's effectiveness in maintaining consistent performance under varying conditions.
}

\subsection{Ablation Studies}

Table~\ref{tab4} presents the ablation study on the proposed self-attention layer. The first row of the table presents the original BERT-base results, and the subsequent rows are the results of the self-attention layer with different types of word-vector sorting strategies.

\begin{table*}[t]
\color{rev-color3}
\caption{Ablation study on the proposed self-attention layer with different self-attention Sort layer \\ implementations strategies on BERT-base \cite{devlin2018bert}.}
\begin{center}
\renewcommand{\arraystretch}{1.39}
\begin{tabular}{lccccccc}
\hline
\textbf{Model$^{\mathrm{a}}$} 
    & \textbf{SST-2}
    & \textbf{MRPC}
    & \textbf{MNLI-M}
    & \textbf{MNLI-MM}
    & \textbf{STS-B}
    & \textbf{IMDB}
    & \textbf{Avg}
    \\ \hhline{========}
    
BERT-base Self-Attention & 93.5 & 88.7 & 84.7 & 83.8 & 84.3 & 93.7 & 88.12 \\ \hline
\quad + Eliminate-Layer & 89.9 & 84.4 & 81.7 & 80.4 & 83.3 & 90.7 & 85.07 \\
\quad + Eliminate-Layer + Random-Sort-Layer & 84.5 & 80.4 & 63.6 & 64.4 & 55.7 & 87.5 & 72.68 \\ 
\quad + Eliminate-Layer + SV-Sort-Layer  & \textbf{92.6} & \textbf{88.0} & \textbf{83.7} & \textbf{82.7} & \textbf{84.2} & \textbf{92.7} & \textbf{87.32} \\ \hline

\multicolumn{8}{l}{$^{\mathrm{a}}$ F1-score for MRPC, Pearson Correlation for STS-B, and Accuracy for the remaining tasks.}
\end{tabular}
\label{tab4}
\end{center}
\end{table*}

\textcolor{rev-color}{
The first approach involves removing the concluding word-vectors from the contextual sequence without the use of a sorting layer. The second approach employs the Random-Sort-Layer, which shuffles word-vectors and randomly eliminates them. The final and better strategy is the proposed strategy that incorporates the SV-Sort-Layer that arranges word-vectors according to their Score values, subsequently removing the concluding word-vectors.
}

As shown in the table, the first strategy shows a significant accuracy drop compared to the original BERT-base and the proposed method. The second strategy has a destructive effect on the model results, and the fine-tuning process becomes completely unstable in CoLA, QQP, QNLI, and RTE datasets not presented in the table. The third strategy is superior to other investigated strategies and proves that the proposed self-attention Sort layer is effective. In these experiments, the hyper-parameters values and the fine-tuning conditions are the same.

\section{Conclusion}
\textcolor{rev-color3}{
This paper presents a latency-adjustable Transformer architecture applicable to a wide range of models, with a marginal decrease in accuracy. We introduce a novel Attention Context Contribution (ACC) metric to evaluate the context contribution of each Transformer encoder layer, demonstrating experimentally that word-vector contributions diminish in the last encoder layers. Extensive experiments on language understanding, text generation, and instruction-tuning tasks validate the suggested technique's effectiveness across diverse datasets, with minimal impact on the input's global context. The paper also provides an accurate analytical estimation of speedup, assisting system designers with hyper-parameter selection. Additionally, the method's powerful offline-tuning capability allows for model speedup adjustments after the fine-tuning phase. The proposed approach highlights that, while the complete network is essential for training, it can be truncated during the fine-tuning phase in Large Language Models (LLMs).
}

\section*{Acknowledgment}
Thanks to Mr. Seyed Mahdi Hosseini for his assistance and comments that greatly improved our work.

\bibliographystyle{IEEEtran}
\bibliography{bibliography.bib}

 
\vspace{11pt}

\begin{IEEEbiography}[{\includegraphics[width=1in,height=1.25in,clip,keepaspectratio]{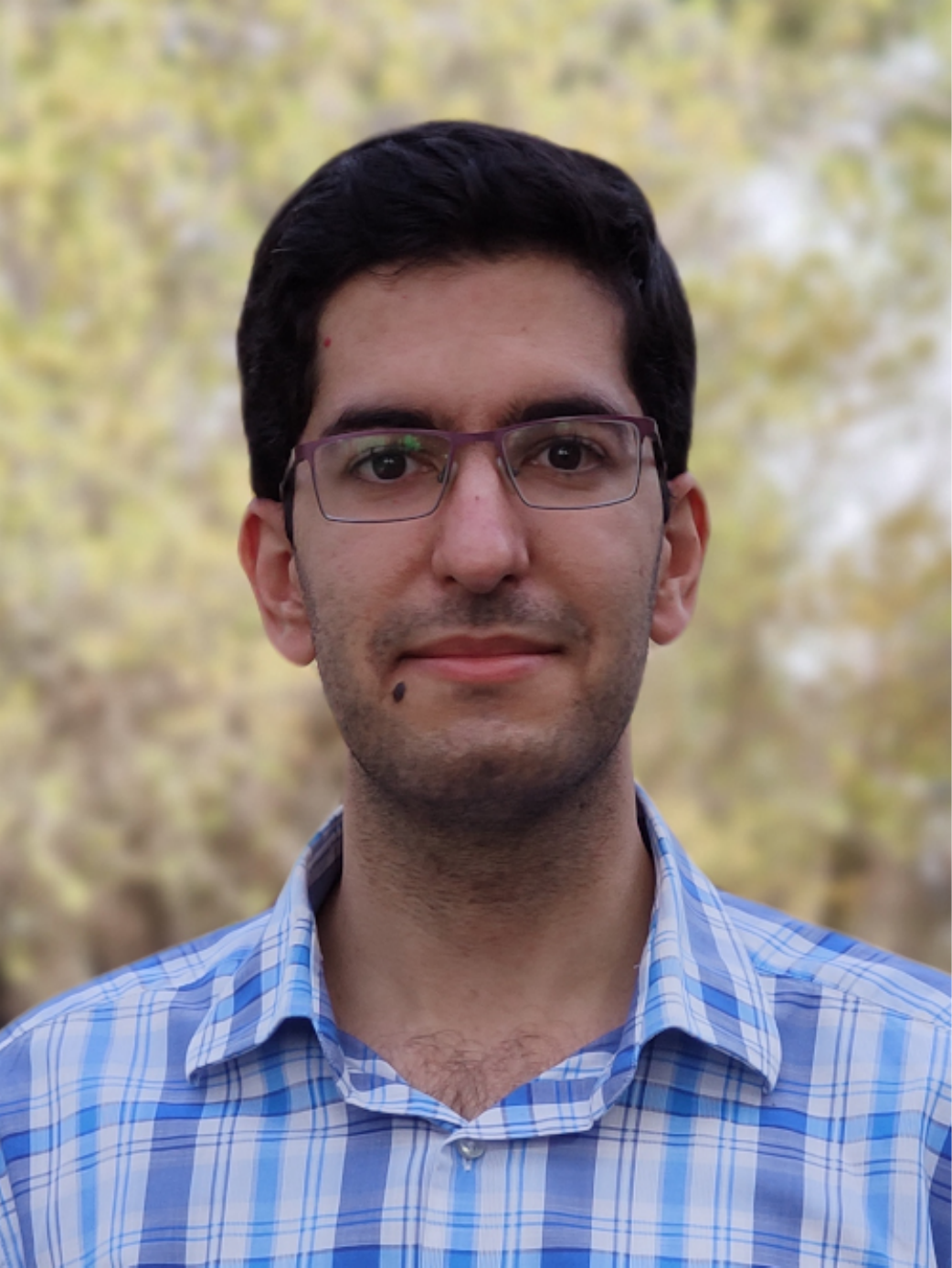}}]{Sajjad Kachuee}
received the B.Sc. and M.Sc. degrees in electrical engineering from Sharif University of Technology, Tehran, Iran, in 2017 and 2019, where he is currently pursuing the Ph.D. degree. His research concerns artificial intelligence and machine learning, focusing on acceleration and optimization of natural language processing (NLP) neural networks.

\end{IEEEbiography}

\vspace{11pt}

\begin{IEEEbiography}[{\includegraphics[width=1in,height=1.25in,clip,keepaspectratio]{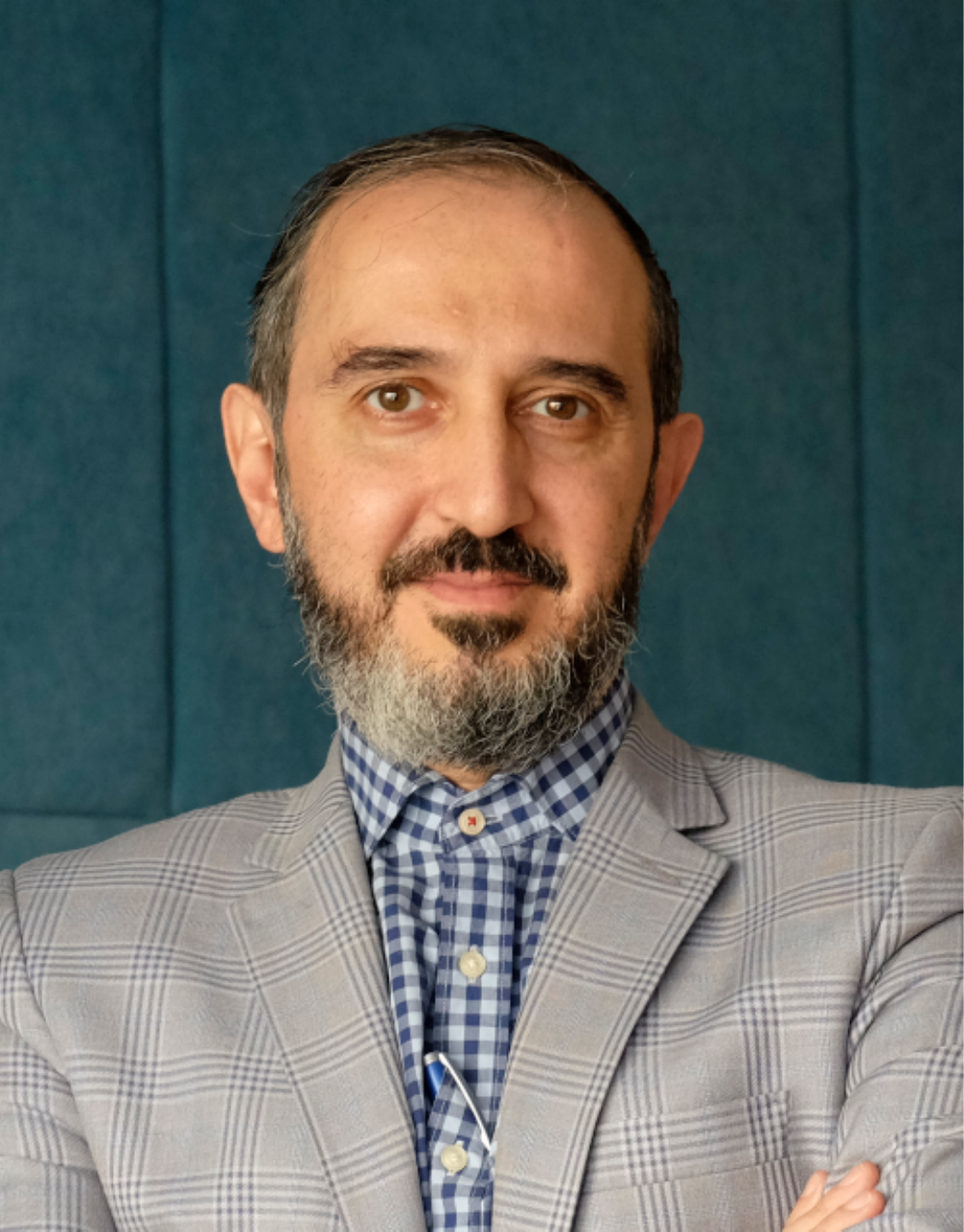}}]{Mohammad Sharifkhani}
received the B.Sc. and M.A.Sc. degrees in electrical and computer engineering from the University of Tehran, Tehran, Iran, in 1998 and 2000, respectively. He received the Ph.D. degree from the University of Waterloo, Waterloo, ON, Canada, in 2006. He was a Postdoctoral Research Fellow at the University of Waterloo in 2007. He is currently an Associate Professor at the Department of Electrical Engineering, Sharif University of Technology, Tehran, Iran. Since 2008 he has published several scientific articles on the broad field of VLSI and Digital Systems. He served as technical committee member and reviewer of several IEEE conferences and journals, respectively. He founded several start-up companies on the broad field of video and image processing as well as machine intelligent systems. His current research is on low-power circuits and architectures, data converters, and application-specific processors for video and machine learning applications.
\end{IEEEbiography}

\vfill

\end{document}